\documentclass[journal]{IEEEtran}
\IEEEoverridecommandlockouts
\usepackage{amsmath,amssymb,amsfonts}
\usepackage{graphicx}
\usepackage{hyperref}
\usepackage[numbers,sort&compress]{natbib}
\usepackage{multirow}
\usepackage{adjustbox}
\usepackage{stfloats}
\usepackage{color}
\usepackage{soul}
\usepackage[inline]{enumitem}
\begin{document}
\bstctlcite{IEEEctl:no_dash}

\title{Benchmarking EMlog Calibration for Autonomous Surface Vehicles}

\author{\IEEEauthorblockN{Samuel Cohen-Salmon\IEEEauthorrefmark{1}, and Itzik Klein}
\IEEEauthorblockA\\{{The Hatter Department of Marine Technologies} \\
{Charney School of Marine Sciences, University of Haifa}\\
Haifa, Israel}

\thanks{\IEEEauthorrefmark{1}Guarantor: Samuel Cohen-Salmon (scohensa@campus.haifa.ac.il).}}

\maketitle

\begin{abstract}
Accurate velocity measurement is a fundamental requirement for autonomous surface and underwater vehicles. Commonly, velocity is provided by a Doppler velocity log (DVL) sensor, yet it becomes unavailable due to operational altitude constraints. In such situations, electromagnetic logs (EMLogs) provide a critically robust alternative for continuous velocity estimation. However, raw EMLog measurements are inherently corrupted by systematic errors, which need to be calibrated prior mission begins.  Currently, a benchmarking comparative evaluation of how different calibration models perform under rapidly changing dynamic sea conditions is missing in the literature. To bridge this gap, this paper presents a comparative model-based calibration methodology that evaluates four distinct calibration models using two different estimation pipelines. The proposed framework is rigorously validated on a unique 221 minutes of continuous real-world telemetry collected from the MARVEL surface vehicle during dynamic sea trials. The dataset contains two different EMLogs and DVL recordings. Experimental results demonstrate that the bias and scale error model implemented with the Kalman filter improves the speed estimation by 71\%. We also demonstrate that dynamical manoeuvres further improve the accuracy compared to standard straight-line paths, ultimately delivering a validated, real-time online calibration EMLog approach for autonomous surface vehicles.
\end{abstract}
\vspace{-6mm}
% Problem Formulation
\section{INTRODUCTION }
\vspace{-1mm}
% General intro to the subject
Autonomous surface vehicles (ASVs) and autonomous underwater vehicles (AUVs) are increasingly critical for a wide spectrum of maritime applications, ranging from oceanographic mapping and environmental monitoring to offshore infrastructure inspection and defense operations~\cite{liu2016unmanned, wynn2014autonomous}. To execute these complex missions safely, these robotic platforms require continuous and highly accurate navigation systems. In global navigation satellite system (GNSS) denied environments, both AUVs and ASVs traditionally rely on Doppler velocity logs (DVLs) as their primary sensor for high-precision velocity measurements. By maintaining an acoustic lock on the seafloor, DVLs enable robust and drift-resistant dead reckoning~\cite{paull2014auv}. For accurate DVL measurements, calibration of the DVL error-terms is required~\cite{yampolsky2025dcnet, wang2019novel, ning2023fast}.

However, DVL bottom-track availability is strictly bounded by the platform's operational altitude relative to the seafloor. For AUVs navigating in deep waters, operating too far from the seabed results in a complete loss of acoustic bottom-track. Conversely, for ASVs operating in shallow coastal areas, passing too close to the seabed causes acoustic interference, signal reverberation, and minimum altitude violations~\cite{kinsey2006survey}. In situations where DVL measurements become unavailable due to altitude constraints and other reasons~\cite{yona2023missbeamnet, klein2022estimating, zhao2019auv}, electromagnetic logs (EMlogs) are critically utilized as a robust alternative to measure the vehicle's speed-through-water or horizontal velocity, ensuring uninterrupted navigation integrity~\cite{ebest2020electromagnetic, alexiou2023sensor}.\\EMlogs are robust maritime velocity sensors that measure a vessel's speed-through-water based on Faraday's law of electromagnetic induction. By generating a magnetic field around the hull, these sensors calculate velocity by measuring the voltage induced by the flow of conductive seawater past the electrodes~\cite{ebest2020electromagnetic}. Due to their reliability, EMlogs are extensively utilized across commercial shipping~\cite{ebest2020electromagnetic} and autonomous marine robotics~\cite{cho2018imm, alexiou2023sensor}. Recent literature frequently integrates these sensors as critical velocity aids, employing them in time-series measurement forecasting~\cite{alexiou2023sensor, zhang2025electro}, and advanced interacting multiple model (IMM) filtering for sea current estimation~\cite{cho2018imm, cha2019integration}.

Despite their ubiquity, raw EMlog measurements are inherently corrupted by systematic errors arising from complex hydrodynamic flow patterns, boundary-layer disturbances, incorrect installation angles, and electronic zero-drift. Rigorous sensor calibration is therefore critically important to mathematically map these uncalibrated readings to true reference speeds, ensuring continuous and accurate navigation for ASVs and AUVs. To address this, various EMLog calibration methodologies have been proposed in the literature. These approaches range from classic post-mission matrix algorithms and static offline bias corrections~\cite{gay1981calibration, brokloff1994matrix} to modern quasi-Newton quaternions and recursive alignment filters designed to isolate and compensate for sensor errors.

While these foundational approaches provide basic error compensation, EMlogs are still largely treated as secondary velocity aids where simple constant biases are either statically calibrated offline or absorbed by high-level navigation filters. The literature currently lacks a systematic comparative evaluation of how different physical error models, from simple constant offsets to non linear hydrodynamic drag, perform against one another under rapidly changing dynamic sea conditions. Specifically, there is limited comparative research between traditional post-processing batch methods~\cite{gay1981calibration} and real-time recursive estimation frameworks~\cite{cho2018imm, zhang2025electro} that mathematically accounts for how specific vehicle maneuver trajectories alter parameter observability and error decoupling.

Despite the widespread use of EMlogs on marine vehicles, the literature on their calibration remains fragmented: existing studies adopt varying error models, ranging from a simple scale factor to more elaborate formulations, and rely on different estimation approaches, making it difficult to compare results across studies or validate them under common conditions. To address this gap, this paper makes the following contributions:
\begin{enumerate}[align=left, leftmargin=*, labelindent=0pt, widest=iii, itemsep=3pt]
\item \textbf{EMlog error models:} Four error models are proposed and investigated. Each proposed error model utilizes a different combination of the EMlog error terms including scale factor and bias.
\item \textbf{Model-based calibration:} Utilizing the four error models we offer seven estimation models in two estimation pipelines:1) least squares to estimate directly the error terms and 2) Kalman filter to estimate the platform speed and error terms.
\end{enumerate}
To evaluate our approach, a unique dataset containing 221 minutes of continuous real-world telemetry collected from the MARVEL ASV, featuring simultaneous measurements from two commercial EMlogs and a high-precision DVL across five distinct dynamic trajectories. This dataset empirically validates our practical, real-time online calibration blueprint, achieving optimal parameter convergence within seconds without the overhead of post-processing.\\
The experimental results demonstrate that the optimal calibration framework reduces speed measurement errors by 71\%. Additionally, high-excitation calibration maneuvers boost accuracy by 30\% compared to standard straight-line paths. As a result of choosing the best combination of the error model and estimation approach, we provide a framework that significantly enhances EMlog calibration performance, allowing for robust, high-fidelity autonomous navigation and dead reckoning in GNSS-denied environments without the overhead of post-processing.

% Structure of the paper
The remainder of this paper is organized as follows: Section II details the mathematical problem formulation for the least squares and Kalman filter frameworks. Section III describes our proposed approach and methodology. Section IV details the dynamic sea trial experimental setup and presents the calibration results, including the cross-validation matrix and spatial trajectory analysis. Finally, Section V concludes the paper with a summary of findings and future directions.
\vspace{-3mm}
\section{PROBLEM FORMULATION}
\vspace{-1mm}
\subsection{Least squares formulation}
\vspace{-.5mm}
The least squares method estimates the optimal parameters of a static model by minimizing the discrepancy between observed data and the mathematical prediction \cite{kay1993fundamentals}. 

Let $\boldsymbol{z} \in \mathbb{R}^{N}$ denote the observation vector containing $N$ continuous measurements, $\mathbf{H_{LS}} \in \mathbb{R}^{N \times p}$ denote the design matrix relating the measurements to the system model, and $p$ is the number of unknown system parameters. The objective is to find the parameter vector $\hat{\boldsymbol{\theta}}_{LS} \in \mathbb{R}^{p}$ that minimizes the sum of squared residuals, defined by the cost function $J(\boldsymbol{\theta})$:
\vspace{-1mm}
\begin{equation}
 J(\boldsymbol{\theta}) = \sum_{k=1}^{N} (z_k - \boldsymbol{h}_{k}^{T} \boldsymbol{\theta})^{2} = ||\boldsymbol{z} - \mathbf{H_{LS}}\boldsymbol{\theta}||^{2} \end{equation}
where $\boldsymbol{\theta} \in \mathbb{R}^{p}$ is the parameter vector being estimated, $z_k$ is the $k$-th scalar measurement in the vector $\boldsymbol{z}$, and $\boldsymbol{h}_{k}^{T}$ represents the $k$-th row vector of the design matrix $\mathbf{H}$.
The optimal analytical solution is derived by solving (1):
%\vspace{-5mm}
\begin{equation} \hat{\boldsymbol{\theta}}_{LS} = (\mathbf{H_{LS}}^T \mathbf{H_{LS}})^{-1} \mathbf{H_{LS}}^T \boldsymbol{Z} \end{equation}
where the matrix $(\mathbf{H_{LS}}^T \mathbf{H_{LS}})$ is assumed to be non-singular to ensure a unique set of estimated parameters.
\vspace{-5mm}
\subsection{Kalman filter}The KF is an optimal recursive algorithm used to estimate the state of a dynamic system from a series of noisy measurements \cite{simon2006optimal, welch1995introduction, brown1997introduction}. The algorithm assumes the system is governed by a linear stochastic difference equation and can be represented in state-space form \cite{cha2019integration, cho2018imm}.The discrete-time linear dynamic system is modeled by the state equation:
\vspace{-1mm}
\begin{equation}
\boldsymbol{x}_k = \boldsymbol{\Phi}_{k-1} \boldsymbol{x}_{k-1} + \boldsymbol{w}_{k-1}
\end{equation}
where $\boldsymbol{x}_k$ represents the system state vector at time step $k$, $\boldsymbol{\Phi}_{k-1}$ is the state transition matrix relating the state from step $k-1$ to $k$, and $\boldsymbol{w}_{k-1}$ is the process noise vector. The process noise is assumed to be drawn from a zero-mean multivariate normal distribution with covariance $\mathbf{Q}$. 
\\The measurement model, relating the true state to the measured variables, is given by:
 \vspace{-1mm}
\begin{equation}
\boldsymbol{z}_k = \mathbf{H}_k \boldsymbol{x}_k + \boldsymbol{v}_k
\end{equation}
\vspace{-5mm}\\
where $\boldsymbol{z}_k$ is the measurement vector, $\mathbf{H}_k$ is the observation matrix, and $\boldsymbol{v}_k$ represents the measurement noise vector. The measurement noise is assumed to be independent of the process noise and drawn from a zero-mean multivariate normal distribution with covariance $\mathbf{R}$. 
\\The recursive estimation operates in two distinct phases: the prediction (time update) and the update (measurement update). In the prediction phase, the filter projects the current state and error covariance forward in time:
\vspace{-1mm}
\begin{equation}\hat{\boldsymbol{x}}_{k|k-1} = \boldsymbol{\Phi}_{k-1} \hat{\boldsymbol{x}}_{k-1|k-1}\end{equation}\begin{equation}\mathbf{P}_{k|k-1} = \boldsymbol{\Phi}_{k-1} \mathbf{P}_{k-1|k-1} \boldsymbol{\Phi}_{k-1}^T + \mathbf{Q}\end{equation}

where $\hat{\boldsymbol{x}}_{k|k-1}$ is the a priori state estimate, $\mathbf{P}_{k|k-1}$ is the a priori estimate error covariance, $\hat{\boldsymbol{x}}_{k-1|k-1}$ is the a posteriori state estimate from the previous step, $\mathbf{P}_{k-1|k-1}$ is the previous a posteriori error covariance, and $\mathbf{Q}$ is the process noise covariance matrix. In the update phase, the filter incorporates the new measurement $\boldsymbol{z}_k$ to refine the state estimate:\begin{equation}\mathbf{K}_k = \mathbf{P}_{k|k-1} \mathbf{H}_k^T (\mathbf{H}_k \mathbf{P}_{k|k-1} \mathbf{H}_k^T + \mathbf{R})^{-1}\end{equation}\begin{equation}\hat{\boldsymbol{x}}_{k|k} = \hat{\boldsymbol{x}}_{k|k-1} + \mathbf{K}_k (\boldsymbol{z}_k - \mathbf{H}_k \hat{\boldsymbol{x}}_{k|k-1})\end{equation}\begin{equation}\mathbf{P}_{k|k} = (\mathbf{I} - \mathbf{K}_k \mathbf{H}_k) \mathbf{P}_{k|k-1}\end{equation}

where $\mathbf{K}_k$ is the optimal Kalman gain (which acts as a blending factor, determining the weight given to the new measurement innovation relative to the a priori estimate), $\hat{\boldsymbol{x}}_{k|k}$ is the updated a posteriori state estimate, $\mathbf{P}_{k|k}$ is the updated a posteriori error covariance, $\mathbf{I}$ is the identity matrix, and the expression $(\boldsymbol{z}_k - \mathbf{H}_k \hat{\boldsymbol{x}}_{k|k-1})$ represents the measurement innovation (or residual).
\vspace{-3mm}
\section{PROPOSED APPROACH}
\vspace{-1mm}
\subsection{Error models} 
\vspace{-1mm}
To accurately characterize the systematic errors of the EMlog, we propose four distinct parametric models. The error models (EM) are:

1) Bias only (EM1): This model includes a constant additive bias, which is typically related to an electronic zero-drift. The measured EMlog speed is expressed as:
\begin{equation}
 \tilde{v}_{k} = v_{GT,k} + b + \eta_k
\end{equation}
where $\tilde{v}_{k} $ is the measured speed, $v_{GT}$ is the ground truth (GT) speed, $b$ is the constant bias parameter and $\eta_k $ is a zero-mean white Gaussian noise.
\vspace{0.5mm}
\\2) Zero offset (EM2): This model assumes zero-offset ($b=0$) but introduces a multiplicative gain error to address hydrodynamic scaling errors or incorrect installation angles. The EM2 speed measurement is:
\vspace{-1mm}
\begin{equation}
 \tilde{v}_{k} = (1+m) \cdot v_{GT,k} + \eta_k
\end{equation} 
where $m$ represents a constant scale factor.
\vspace{0.5mm}
\\3) Bias and scale model (EM3): Represents the general linear case by combining both scale and bias errors. It is the most common approximation for marine speed logs, as it captures both hydrodynamic gain errors and electronic or installation offsets simultaneously \cite{gay1981calibration, zhang2025electro}:
\begin{equation}
 \tilde{v}_{k} = (1+m) \cdot v_{GT,k} + b + \eta_k
\end{equation}

4) Polynomial model (EM4): Expands the linear affine framework into a second-order polynomial to account for non linear hydrodynamic effects such as boundary layer flow separation and hull curvature disturbances, which must be compensated for during electromagnetic log calibration \cite{fluids6020066,dcbfa405d31e4af28904054dcda55318, gay1981calibration}:
\begin{equation}
 \tilde{v}_{k} = a\cdot v_{GT,k}^2 + (1+m) \cdot v_{GT,k} + b + \eta_k
\end{equation} 

where $a$ represents the quadratic hydrodynamic coefficient.\vspace{0.5mm} 

Having mathematically defined the four discrete parametric representations of the sensor's error profile, the operational challenge shifts to configuring the estimator blocks to extract these coefficients. 
\vspace{-4mm}
\subsection{Parameter Identifiability Analysis}
Drawing on standard parameter identifiability principles from estimation theory~\cite{kay1993fundamentals}, which have been actively adapted for marine velocity sensor observability~\cite{fossen2011handbook}, we evaluate the mathematical condition required to decouple the scale factor $m$ from the zero-drift bias $b$. Focusing on the baseline linear affine model (EM3) defined in (20), structural identifiability is governed by the $2 \times 2$ information matrix product $\boldsymbol{H}^T \boldsymbol{H}$ derived from the design matrix $\boldsymbol{H} \in \mathbb{R}^{N \times 2}$:
\begin{equation}
\boldsymbol{H}^T \boldsymbol{H} = \begin{bmatrix} \sum_{k=1}^{N} v_{GT,k}^2 & \sum_{k=1}^{N} v_{GT,k} \\ \sum_{k=1}^{N} v_{GT,k} & N \end{bmatrix}
\end{equation}

When the vehicle maintains a constant forward cruise speed ($v_{GT,k} = V_c$), substituting this steady velocity constraint into $\boldsymbol{H}^T \boldsymbol{H}$ simplifies the information matrix to:
\begin{equation}
\boldsymbol{H}^T \boldsymbol{H} = \begin{bmatrix} N V_c^2 & N V_c \\ N V_c & N \end{bmatrix}
\end{equation}

Evaluating the determinant yields:
\begin{equation}
\begin{aligned}
\det(\boldsymbol{H}^T \boldsymbol{H}) &= (N V_c^2)(N) - (N V_c)^2 
\\&= N^2 V_c^2 - N^2 V_c^2 = 0
\end{aligned}
\end{equation}
\vspace{-3mm}\\
Because $\det(\boldsymbol{H}^T \boldsymbol{H}) = 0$, the information matrix is strictly singular and rank-deficient (\mbox{$\text{rank}(\boldsymbol{H}^T \boldsymbol{H}) = 1$}). While this proof explicitly addresses the linear affine framework, the identical structural singularity applies to the polynomial formulation defined in (21). Under constant speed, the quadratic term $v_{GT,k}^2$ simply becomes another collinear constant $V_c^2$, fundamentally collapsing the expanded $3 \times 3$ information matrix to a rank of 1.

Conversely, executing dynamic maneuvers introduces speed fluctuations over time, satisfying the structural non-collinearity condition:
\vspace{-2mm}
\begin{equation}
\begin{aligned}
\left(\sum_{k=1}^{N} v_{GT,k}^2 \right) \cdot N \neq \left(\sum_{k=1}^{N} v_{GT,k}\right)^2 \\\implies \det(\boldsymbol{H}^T \boldsymbol{H}) > 0
\end{aligned}
\end{equation}

This inequality guarantees that \mbox{$\det(\boldsymbol{H}^T \boldsymbol{H}) > 0$}, raising the information matrix to full rank \mbox{($\text{rank}(\boldsymbol{H}^T \boldsymbol{H}) = 2$)} and decoupling the scale factor $m$ from the zero-drift bias $b$.
\vspace{-4mm}
\subsection{Methodology}
We suggest and define four least squares variations and three Kalman filters variations across the proposed error models, aiming to identify the most appropriate combination of error representation and estimation framework.

\subsubsection{Least squares calibration}
For the batch LS implementations, we construct specific design matrices ($\mathbf{H}$) and parameter vectors ($\boldsymbol{\theta}$) based on the normal equations to target each error profile.
\begin{enumerate}[label=\roman*), align=left, leftmargin=*, labelindent=0pt, widest=iii]
\item LS-EM1: The design matrix is constructed from (10) as a column vector of ones, and the parameter isolates the bias as \vspace{-2mm}\begin{equation}
 \mathbf{H} = [1 \dots 1]^T, \quad \boldsymbol{\theta} = [b]
\end{equation}

\item LS-EM2: Following (11), the design matrix consists of the uncalibrated speed measurements to estimate the scale factor, 
\vspace{-3mm}
 \begin{equation}
 \mathbf{H} = \begin{bmatrix}
     v_{GT,1} & \dots & v_{GT, N}
 \end{bmatrix}^T, \quad \boldsymbol{\theta} = [m]
\end{equation}
\item LS-EM3: This model combines (18) and (19) by:
\begin{equation}
 \mathbf{H} = \begin{bmatrix}
v_{GT,1} & \dots & v_{GT,N} \\
1 & \dots & 1
\end{bmatrix}^T, \quad \boldsymbol{\theta} = [m \quad b]^T
\end{equation}
\item LS-EM4: The polynomial LS model expands the design vector to capture non linearities, yielding:
\begin{equation}
\begin{aligned}
 \mathbf{H} = \begin{bmatrix}
 v_{GT,1}^2 & \dots & v_{GT,N}^2 \\
 v_{GT,1} & \dots & v_{GT,N} \\
 1 & \dots & 1
 \end{bmatrix}^T,
 \boldsymbol{\theta} &= [a \quad m \quad b]^T
\end{aligned}
\end{equation}
\end{enumerate}
\subsubsection{Kalman filter calibration} 
For the KF implementations, we construct specific state vectors ($\boldsymbol{x}_k$) and time-varying observation matrices ($\boldsymbol{H}_k$) to estimate the scalar speed ($v_k$) simultaneously with the unknown sensor error parameters. To preserve strict system linearity without requiring an extended Kalman filter (EKF), a pseudo-linear operating point approach is adopted where the GT reference speed $v_{GT,k}$ populates the observation matrix $\boldsymbol{H}_k$. Across all variants, the state transition matrix is defined as an identity matrix $\boldsymbol{\Phi}_k = \mathbf{I}_{p \times p}$ matching state vector dimension $p$, assuming static or random walk error states.
\\We define three different models for the KF calibration tasks:
\begin{enumerate}[label=\roman*), align=left, leftmargin=*, labelindent=0pt, widest=iii]
\item KF-EM1: A two-dimensional state vector, $\boldsymbol{x}_k$, that includes the scalar speed and constant bias, with a corresponding measurement matrix $\boldsymbol{H}_k$.\vspace{-1mm}
 \begin{equation}
 \boldsymbol{x}_k = \begin{bmatrix} v_k \\ b_k \end{bmatrix}, \quad \boldsymbol{H}_k = \begin{bmatrix} 1 & 1 \end{bmatrix}
 \end{equation}\vspace{-3mm}
\item KF-EM2: A two-dimensional state vector, $\boldsymbol{x}_k$, that includes the scalar speed and multiplicative scale factor, with a corresponding time-varying measurement matrix $\boldsymbol{H}_k$.
\vspace{-1mm}
 \begin{equation}
 \boldsymbol{x}_k = \begin{bmatrix} v_k \\ m_k \end{bmatrix}, \quad \boldsymbol{H}_k = \begin{bmatrix} 1 & v_{GT,k} \end{bmatrix}
 \end{equation}
\item KF-EM3: A three-dimensional state vector, $\boldsymbol{x}_k$, that includes the scalar speed, scale factor, and zero-drift bias, with a corresponding time-varying measurement matrix $\boldsymbol{H}_k$.
 \begin{equation}
 \boldsymbol{x}_k = \begin{bmatrix} v_k \\ m_k \\ b_k \end{bmatrix}, \quad \boldsymbol{H}_k = \begin{bmatrix} 1 & v_{GT,k} & 1 \end{bmatrix}
 \end{equation}
\end{enumerate}
\subsubsection{Summary}
\begin{figure}[!h]
	\centering
 \vspace{-4mm}
	\includegraphics[width=\columnwidth]{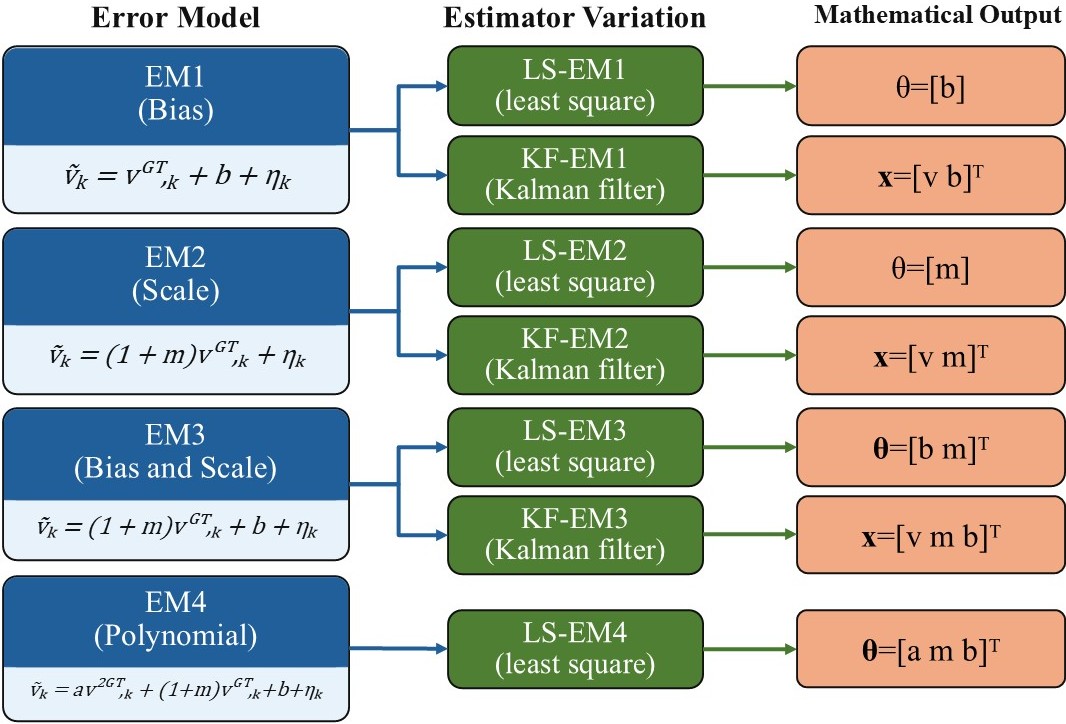}
 \vspace{-8mm}
	 \caption{Overview of the proposed EMlog calibration architecture. The framework maps four distinct parametric error models (left) to seven parallel estimator variations (center) utilizing both least squares and Kalman filtering. The corresponding mathematical output vectors (right) detail the extraction of calibration parameters ($\theta$) dynamic states ($x_k$).}\label{model_fig}
\end{figure}

In total, we examined seven calibration methodologies, consisting of four LS variations and three KF variations. The estimator architectures transition from targeting single-parameter disturbances (isolated bias or scale factor) to estimating their coupled linear affine representation and higher-order hydrodynamic interactions. A visual overview of the examined models and their affiliation is presented in Fig.\ref{model_fig}.

\vspace{-6mm}
% Results
\section{ANALYSIS AND RESULTS}\label{res_sec}
\begin{figure}[!h]
	\centering
 \vspace{-3mm}
	\includegraphics[width=\columnwidth]{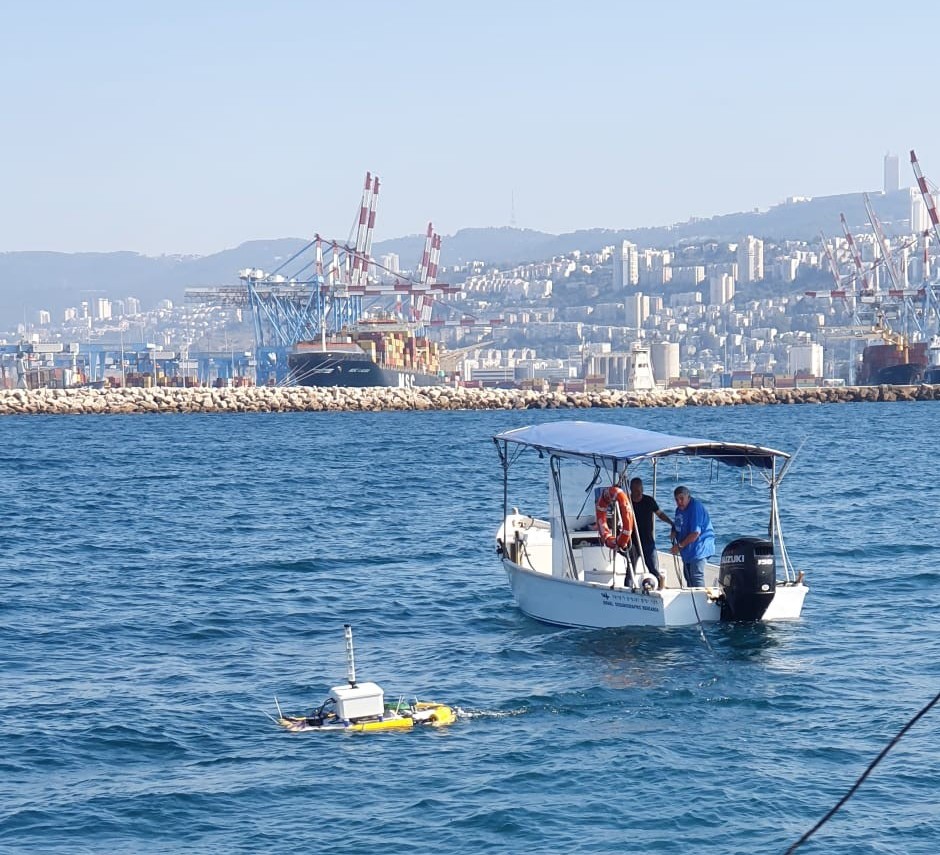}
 \vspace{-6mm}
	 \caption{The MARVEL platform and its safety boat deployed during dynamic sea trials for sensor data collection.}\label{marvel_fig}
 \vspace{-7mm}
\end{figure}

To evaluate the proposed model-based calibration framework, this section presents a comprehensive empirical validation using real-world experimental data collected from the MARVEL ASV during dynamic sea trials, as shown in Fig. \ref{marvel_fig}. We first detail the experimental setup and hardware architecture, followed by the recorded trajectory dataset, preprocessing pipeline, and temporal synchronization methods. Next, the evaluation metrics are defined to quantify speed estimation accuracy and estimator consistency. Finally, we present the performance results across all seven estimation models.

\vspace{-2mm}
\subsection{Experimental setup}
\vspace{-1mm}
To perform a rigorous empirical evaluation of the proposed calibration frameworks, dynamic sea trials were executed using dedicated maritime hardware and high-precision reference instrumentation. The physical configuration, sensor specifications, and environmental baseline are structured as follows: the experimental data was collected using the MARVEL ASV as the primary platform for sensor mounting and data acquisition during dynamic maneuvers \cite{ocean2025}. A dedicated safety vessel accompanied the ASV throughout the trials to monitor telemetry and maintain operational safety. 
\\Two distinct commercial electromagnetic speed logs were deployed simultaneously to evaluate performance across different sensor architectures: 
\begin{enumerate}[itemsep=3pt]
\item EMlog1 (airmar DX900+)~\cite{airmar_dx900}: A dual-axis electromagnetic sensor providing longitudinal and transverse speed through water ($V_{\text{HW}}$) at a 1Hz update rate. The manufacturer-specified accuracy is $\pm 0.1$ knots for speeds below 10 knots and $\pm 1\%$ for speeds exceeding 10 knots. 
\item EMlog2 (ben marine alize)~\cite{benmarine_alize}: An electromagnetic speed log outputting 2D water speed ($V_{\text{BW}}$) and pitch/roll data at 10Hz, alongside 1D water speed ($V_{\text{HW}}$) and current angle at 5HZ. The system specifies an operational accuracy better than 0.5 knots post-calibration. 
\end{enumerate}
To evaluate the primary unit under test (UUT) here EMlog1, two distinct GT velocity references were utilized: 1) the secondary EMlog sensor (EMlog2) for relative inter-sensor calibration, and 2) a high-precision Rowe SeaPILOT DVL~\cite{rowe_seapilot} for absolute velocity benchmarking. Operating at 10\,Hz, the DVL delivers high-resolution velocity measurements with a long-term accuracy of $\pm 0.25\%$ and a speed resolution of $0.01\text{ cm/s}$.
\\To isolate sensor dynamics from ambient environmental disturbances, regional oceanographic data was cross-referenced with the Israel oceanographic and limnological research (IOLR) meteo-marine database~\cite{iolr_data}. The database confirmed that ambient sea currents during the trial window were negligible, reaching a maximum value of $0.286\text{ m/s}$. This minimal current activity physically justifies treating the DVL bottom-track velocity as an absolute GT reference for water-relative speed calibration~\cite{iolr_data}.\vspace{-5mm} 
\subsection{Dataset and preprocessing}
\vspace{-1mm}
To validate the calibration framework under realistic operational conditions, the raw telemetry recorded during sea trials underwent trajectory segmentation, temporal alignment, and a systematic combinatorial cross-validation protocol, as described below:
\subsubsection{Dataset}
\begin{figure}[!h]
	\centering
 \vspace{-3mm}
	\includegraphics[width=\columnwidth]{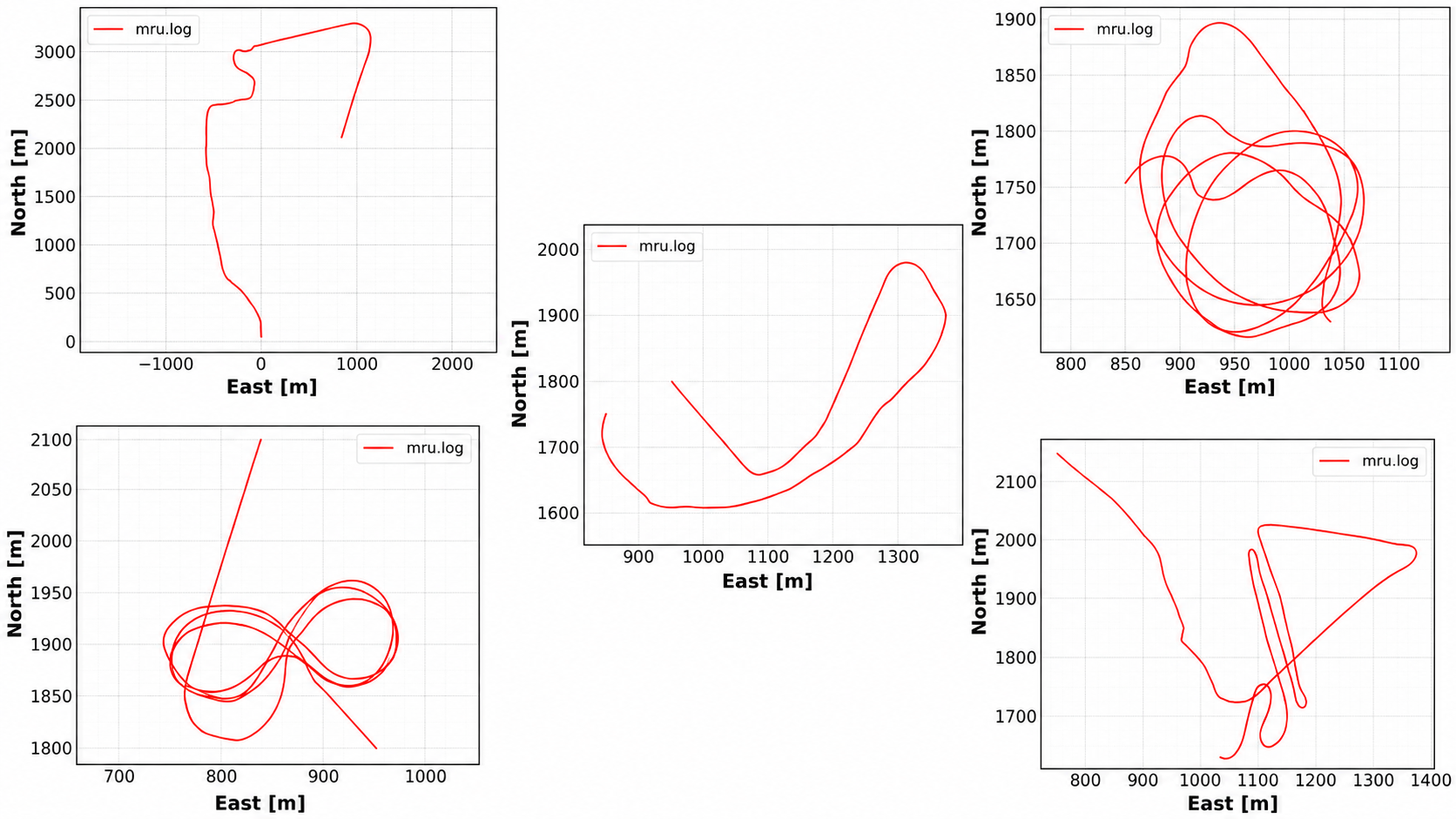}
 \vspace{-7mm}
	 \caption{2D Spatial trajectories (East vs. North) of the ASV across the five segmented trial sections:transit to experiment location (top-left), circular maneuvers (top-right), clean/straight trajectories (center), figure-8 maneuvers (bottom-left), and transit back (bottom-right).}\label{calib_traj_fig}
 \vspace{-3mm}
\end{figure}

The continuous operational dataset was segmented into five distinct trajectories, each exhibiting unique dynamic excitation characteristics, durations, and vessel speed profiles:
\begin{itemize}
\item Trajectory 1: (transit to experiment location): Duration of 125 min with an average forward speed of 1.95 m/s.
\item Trajectory 2: (figure-8 maneuvers): Duration of 27 min (roughly 1.5 minute per cycle) with an average forward speed of 1.48 m/s. This segment introduces transient speed fluctuations, variable slip angles, and alternating port/starboard turn dynamics. 
\item Trajectory 3: (clean / straight lines): Duration of 15 min with an average forward speed of 1.59 m/s. This trajectory represents steady-state, non-accelerating forward transits.
\item Trajectory 4: (circular maneuvers): Duration of 24 min with an average forward speed of 1.85 m/s. This segment drives the vehicle through continuous $360^\circ$ loops, varying heading relative to environmental forces. 
\item Trajectory 5: (transit back): Duration of 30 min with an average forward speed of 1.37 m/s. Due to the main batteries running out, only the first 30 minutes of this segment were recorded (out of 125 as in Trajectory 1).
\end{itemize}
In total the dataset contains 221 minutes of recorded data from both EMlogs and DVL. Fig. \ref{calib_traj_fig} presents the horizontal position of the five trajectories recorded using MARVEL.

\subsubsection{Implementation details}
In all KF implementations, the state vector is initialized with the physical assumption of an ideal sensor, setting initial scale factor $m_0 = 0$ and bias $b_0 = 0$. The initial error covariance matrix $\boldsymbol{P}_0$ is set conservatively to $0.25 \cdot \mathbf{I}$, and the process noise covariance matrix $\boldsymbol{Q}$ is assigned a nominal non-zero value of $10^{-7} \cdot \mathbf{I}$ to prevent covariance collapse and ensure continuous estimator responsiveness \cite{zarchan2015polynomial, brown1997introduction}. The measurement noise variance ($\boldsymbol{R}$) is initialized based on the sensor's specifications \cite{simon2006optimal}.
\subsubsection{Temporal synchronization and lag optimization} Because the EMlog1 updates at 1HZ while the EMlog2 and DVL operate at 10HZ, raw measurements exhibit sampling rate asynchronous delays and phase shifts during dynamic maneuvers. To eliminate data skewing without generating false filter innovations, a cross-correlation optimization algorithm was executed over the initialization phase. The temporal alignment objective maximizes the cross-correlation coefficient $R_{xy}(\tau)$ between the measured speed magnitude $v_{\text{meas}}(t)$ and GT speed $v_{\text{ref}}(t)$ over discrete time shift $\tau$:
\begin{equation}
 \hat{\tau} = \arg\max_{\tau} \sum_{t} v_{\text{meas}}(t+\tau) \cdot v_{\text{ref}}(t)
\end{equation}
This optimization resolved transmission and serial communication delays of $\hat{\tau}_1 = 3.2\text{ s}$ for EMlog1 and $\hat{\tau}_2 = 0.45\text{ s}$ for EMlog2. Applying these latency shifts ensured precise alignment across all channels prior to estimator ingestion. 
\vspace{-3mm} 
\subsection{Evaluation metric}
To quantitatively assess speed tracking accuracy and evaluate recursive filter stability across all tested error model variations, two primary performance metrics are defined:
\subsubsection{Root mean square error} The primary performance benchmark for speed correction accuracy across the seven estimation models is the speed RMSE:
\begin{equation}
 \text{RMSE} = \sqrt{\frac{1}{N}\sum_{k=1}^{N}\left(v_{\text{ref},k} - \hat{v}_{\text{cal},k}\right)^2}
\end{equation}
where $v_{\text{ref},k}$ is the GT DVL reference speed at sample step $k$, $\hat{v}_{\text{cal},k}$ is the corrected speed of the UUT computed using the estimated model parameters, and $N$ is the total number of discrete measurements in the evaluated sequence.

\subsubsection{Filter consistency and normalized innovation squared} 
%\vspace{-1mm}
To ensure that the KF remain statistically stable and do not become overconfident, filter consistency is rigorously validated using the normalized innovation squared (NIS) metric, which evaluates consistency directly through the observable measurements\cite{bar2001estimation}. The measurement residual (innovation) $y_k$ at step $k$ is defined as:\vspace{-2mm}
\begin{equation}
 y_k = z_k - \boldsymbol{H}_k \hat{x}_{k\vert{}k-1}
\end{equation}
The innovation covariance matrix $\boldsymbol{S}_k$ is:
\begin{equation}
 \boldsymbol{S}_k = \boldsymbol{H}_k \boldsymbol{P}_{k\vert{}k-1} \boldsymbol{H}_k^T + \boldsymbol{R}
\end{equation}
For a scalar speed measurement the NIS is :
\begin{equation}
 \text{NIS}_k =y_k^{T}S_{k}^{-1}y_k= \frac{y^2_k}{S_k}
\end{equation}
By verifying that the NIS fluctuates appropriately around the expected level, we confirm that the filter covariance accurately bounds the actual estimation errors.
\begin{table*}[b]
\centering
\caption{RMSE and improvement - calibration on figure-8 Run (applied to trajectory 3)}
\label{rmse_improvement_table_2}
\renewcommand{\arraystretch}{1.3} 
\resizebox{\textwidth}{!}{% 
\begin{tabular}{|l|c|cc|cc|cc|cc|cc|cc|cc|}
\hline
\multirow{3}{*}{\textbf{UUT vs GT}} & \textbf{RAW} & \multicolumn{8}{c|}{\textbf{Least square}} & \multicolumn{6}{c|}{\textbf{Kalman filter}} \\ \cline{3-16} 
 & \textbf{RMSE} & \multicolumn{2}{c|}{\textbf{LS-EM1}} & \multicolumn{2}{c|}{\textbf{LS-EM2}} & \multicolumn{2}{c|}{\textbf{LS-EM3}} & \multicolumn{2}{c|}{\textbf{LS-EM4}} & \multicolumn{2}{c|}{\textbf{KF-EM1}} & \multicolumn{2}{c|}{\textbf{KF-EM2}} & \multicolumn{2}{c|}{\textbf{KF-EM3}} \\ \cline{3-16} 
 & \textbf{[m/s]} & \textbf{RMSE} [m/s] & \textbf{\%} & \textbf{RMSE} [m/s] & \textbf{\%} & \textbf{RMSE} [m/s] & \textbf{\%} & \textbf{RMSE} [m/s] & \textbf{\%} & \textbf{RMSE} [m/s] & \textbf{\%} & \textbf{RMSE} [m/s] & \textbf{\%} & \textbf{RMSE} [m/s] & \textbf{\%} \\ \hline
EMlog1 vs DVL & 0.576 & 0.208 & 64\% & 0.276 & 52\% & 0.167 & 71\% & 0.172 & 70\% & 0.208 & 64\% & 0.275 & 52\% & 0.170 & 70\% \\ \hline
EMlog2 vs DVL & 0.305 & 0.113 & 63\% & 0.142 & 53\% & 0.103 & 66\% & 0.104 & 66\% & 0.113 & 63\% & 0.143 & 53\% & 0.102 & 67\% \\ \hline
EMlog1 vs EMlog2 & 0.309 & 0.195 & 37\% & 0.219 & 29\% & 0.177 & 43\% & 0.177 & 43\% & 0.199 & 35\% & 0.221 & 28\% & 0.182 & 41\% \\ \hline
\end{tabular}%
}
\end{table*}
\subsubsection{Combinatorial cross-validation paradigm}
\begin{figure}[!h]
	\centering
 \includegraphics[width=\columnwidth]{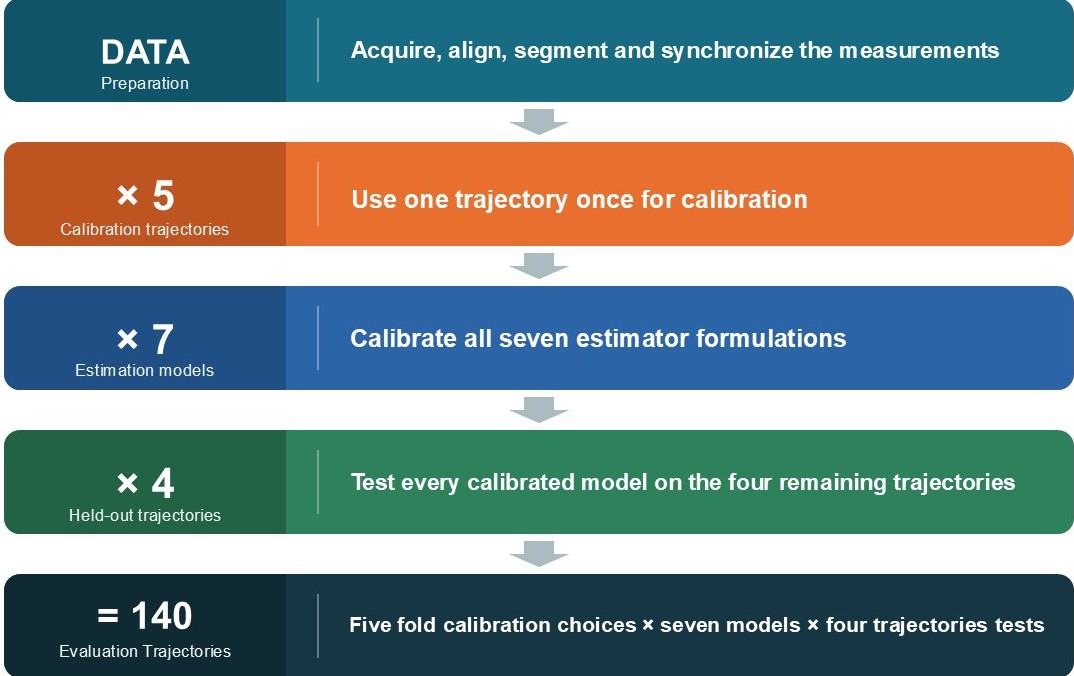}
	 \caption{Block diagram of the proposed EMlog calibration framework. The processing pipeline illustrates the complete workflow}\label{block_diagram_fig}
 \vspace{-6mm}
\end{figure}

To rigorously quantify generalization capability, a five fold cross-validation scheme was implemented across both batch and recursive estimation methods. Each of the five trajectories was independently isolated as a training segment. For LS formulations, optimal parameter vectors ($\boldsymbol{\hat{\theta}}_{\text{LS}}$) were calculated over the training segment via batch inversion. For KF formulations, the estimator was processed over the training segment until the parameter state vector ($\boldsymbol{x}_k$) reached steady-state convergence. For both frameworks, the derived parameter estimates were subsequently frozen and evaluated against the remaining four unseen testing segments to compute out-of-sample speed residuals. In each fold, one of the five trajectories served as the calibration dataset for all seven estimator formulations. The resulting parameter estimates were then frozen and applied separately to each of the four remaining trajectories. For each evaluation, the RMSE between the calibrated UUT and the reference unit was compared with the corresponding RMSE before calibration, and the relative reduction in RMSE was reported as the calibration improvement. This complete sequence of pipeline building blocks is illustrated in Fig. \ref{block_diagram_fig}.\vspace{-1mm}

%\vspace{-4mm}
\subsection{Results}
Due to the exhaustive nature of the 140 combinatorial cross-validation matrix, presenting every individual result is impractical. Therefore, the empirical findings highlight the optimal and most representative configurations. Specifically, Table~\ref{rmse_improvement_table_2} details a representative dynamic scenario (trained on Figure-8, evaluated on straight transits) to benchmark the comparative performance of all seven estimator variations. Conversely, Table~\ref{cv_matrix_table} expands upon the best-performing error model (EM3) to demonstrate its spatial generalization capabilities across all trajectory combinations. Across the entire 140-run matrix, both LS-EM3 and KF-EM3 implementations consistently yielded the highest accuracy, justifying their selection as the primary focal points for the spatial analysis.\\
To evaluate the seven parallel estimator formulations, we applied the segmented trial datasets (illustrated in Fig.~\ref{calib_traj_fig}) to the proposed processing pipeline(Fig.~\ref{block_diagram_fig}). The empirical results are presented in three continuous phases: least squares comparative performance, Kalman filter tracking, and spatial cross-validation insights.

The empirical results presented in Table~\ref{rmse_improvement_table_2} highlight significant performance distinctions across the evaluated error formulations when trained on figure-8 dynamics (trajectory 2) and evaluated on clean straight transits (trajectory 3). 
\\ Evaluating the batch least squares implementations, the baseline constant offset model (\text{LS-EM1}) reduced the raw \text{EMlog1} RMSE from $0.576\text{ m/s}$ to $0.208\text{ m/s}$ ($64\%$ improvement), while the scale factor model (\text{LS-EM2}) yielded a $52\%$ error reduction ($0.276\text{ m/s}$). The combined linear affine formulation (\text{LS-EM3}) achieved the highest calibration accuracy, reducing \text{EMlog1} RMSE down to $0.167\text{ m/s}$ a $71\%$ overall error reduction. Similar performance gains were recorded for \text{EMlog2}, improving RMSE from $0.305\text{ m/s}$ to $0.103\text{ m/s}$ ($66\%$ optimization).\\ For relative calibration between the two speed logs (\text{EMlog1} vs. \text{EMlog2}) without relying on a DVL reference, the linear affine model (\text{LS-EM3}) reduced the raw differential RMSE from $0.309\text{ m/s}$ down to $0.177\text{ m/s}$ ($43\%$ improvement). Finally, expanding to a second-order polynomial (\text{LS-EM4}) yielded an RMSE of $0.172\text{ m/s}$ ($70\%$ improvement), offering no performance gain over the linear affine model and indicating that non linear hydrodynamic effects are negligible within the operational speed envelope.
\begin{table*}[b]
\centering
\caption{Cross-validation :EMlog1 vs DVL calibration and processed trajectories}
\label{cv_matrix_table}
\renewcommand{\arraystretch}{1.3}
\resizebox{\textwidth}{!}{%
\begin{tabular}{|c|c|c|cc|cc|cc|cc|cc|cc|}
\hline
\multirow{3}{*}{\textbf{\begin{tabular}[c]{@{}c@{}}Calibration\\ trajectory\end{tabular}}} & \multirow{3}{*}{\textbf{\begin{tabular}[c]{@{}c@{}}Duration\\ {[}min{]}\end{tabular}}} & \multirow{3}{*}{\textbf{\begin{tabular}[c]{@{}c@{}}Avg. speed\\ {[}m/s{]}\end{tabular}}} & \multicolumn{12}{c|}{\textbf{trajectory}} \\ \cline{4-15}
 & & & \multicolumn{2}{c|}{\textbf{1 (transit)}} & \multicolumn{2}{c|}{\textbf{2 (figure-8)}} & \multicolumn{2}{c|}{\textbf{3 (straight)}} & \multicolumn{2}{c|}{\textbf{4 (circles)}} & \multicolumn{2}{c|}{\textbf{5 (return)}} & \multicolumn{2}{c|}{\textbf{overall}} \\ \cline{4-15}
 & & & \textbf{RMSE [m/s]} & \textbf{Imp [\%]} & \textbf{RMSE [m/s]} & \textbf{Imp [\%]} & \textbf{RMSE [m/s]} & \textbf{Imp [\%]} & \textbf{RMSE [m/s]} & \textbf{Imp [\%]} & \textbf{RMSE [m/s]} & \textbf{Imp [\%]} & \textbf{RMSE [m/s]} & \textbf{Imp [\%]} \\ \hline
\textbf{1 (transit)} & 125 & 1.95 & 0.264 & 54\% & 0.233 & 59\% & 0.191 & 67\% & 0.240 & 40\% & 0.388 & 27\% & 0.273 & 50\% \\ \hline
\textbf{2 (figure-8)} & 27 & 1.48 & 0.341 & 40\% & 0.185 & 67\% & 0.167 & 71\% & 0.271 & 32\% & 0.329 & 38\% & 0.291 & 46\% \\ \hline
\textbf{3 (straight)} & 15 & 1.59 & 0.487 & 15\% & 0.210 & 63\% & 0.147 & 74\% & 0.507 & -27\% & 0.452 & 14\% & 0.423 & 22\% \\ \hline
\textbf{4 (circles)} & 24 & 1.85 & 0.309 & 46\% & 0.205 & 64\% & 0.209 & 64\% & 0.165 & 59\% & 0.283 & 46\% & 0.261 & 52\% \\ \hline
\textbf{5 (return)} & 30 & 1.37 & 0.331 & 42\% & 0.221 & 61\% & 0.231 & 60\% & 0.175 & 56\% & 0.280 & 47\% & 0.277 & 49\% \\ \hline
\end{tabular}%
}
\end{table*}

\begin{figure}[!ht]
 \raggedright
 \includegraphics[width=\columnwidth]{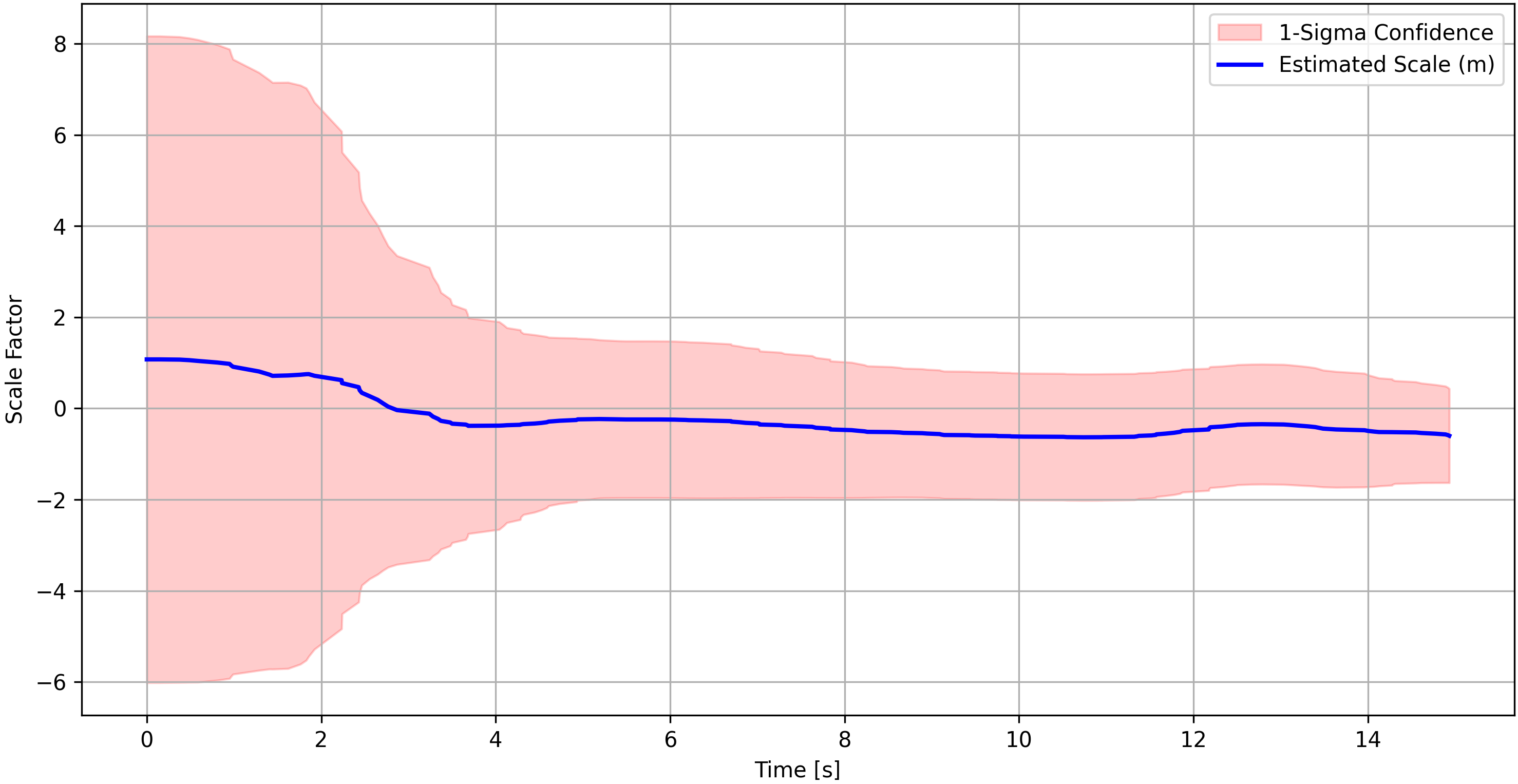}
 \vspace{-8mm}
	 \caption{Scale factor convergence sleeve demonstrating in the KF-EM3 model.}\label{funnel_fig}
     \vspace{-4mm}
\end{figure}

\begin{figure}[!h]
	\centering
	\includegraphics[width=\columnwidth]{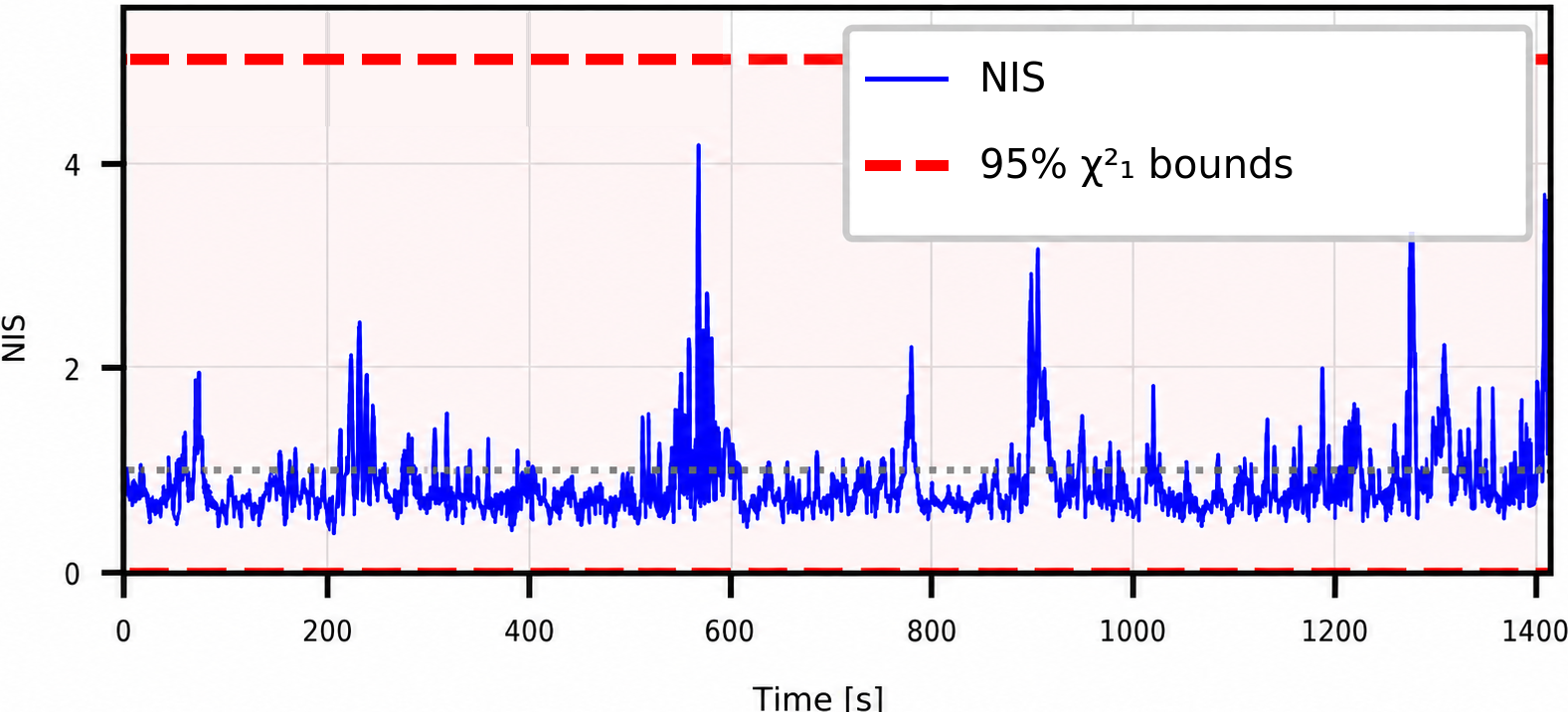}
 \vspace{-7mm}
    \caption{Filter consistency check over the evaluation trajectory. The computed NIS metric is plotted against the expected theoretical $\chi^2$ mean for a single degree of freedom ($\text{NIS} = 1$).}\label{consistency_fig}
 \vspace{-5mm}
\end{figure}

\begin{figure}[!h]
 \centering
 \includegraphics[width=\columnwidth]{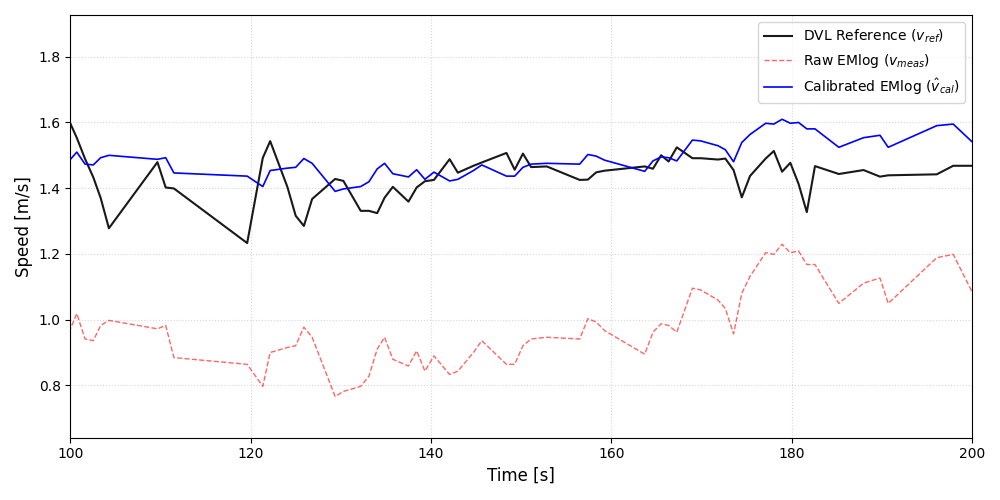}
 \vspace{-9mm}
 \caption{Speed comparison over a 100-second window of the clean trajectory segment (trajectory 3). The plot demonstrates the high correlation between the ground truth DVL reference (black) and the calibrated EMlog outputs (blue) compared to the uncalibrated raw measurements (red dashed).}\label{vel_comp_fig}
 \vspace{-7mm}
\end{figure}

Transitioning to real-time execution, the Kalman filter implementations demonstrated rapid convergence, robust statistical stability, and near-identical performance to the LS methods during dynamic trial segments. For instance, the \text{KF-EM3} formulation similarly achieved a $0.170\text{ m/s}$ RMSE ($70\%$ improvement) for \text{EMlog1} and a $0.182\text{ m/s}$ RMSE ($41\%$ improvement) for inter-sensor relative calibration. 
\\Fig. \ref{funnel_fig} shows the convergence sleeve of the scale-factor in the KF-EM3 model. Filter consistency was strictly maintained; the NIS analysis, depicted in Fig.~\ref{consistency_fig}, confirms that the estimator remains statistically consistent. The computed NIS values fluctuate appropriately around the expected level ($\text{NIS} = 1$) across the trial duration, proving robust statistical stability. Ultimately, the visual comparison Fig.~\ref{vel_comp_fig} demonstrates excellent spatial alignment between the calibrated \text{EMlog} output and the GT DVL velocity benchmark during steady-state transits.

The exhaustive $7 \times 5 \times 4 = 140$ (\ref{block_diagram_fig}) evaluation for LS-EM3 is summarized in Table~\ref{cv_matrix_table}. It reveals a critical dependency on calibration path geometry, offering practical trajectory insights. Parameters trained on high-excitation trajectories, specifically circular maneuvers (trajectory 4) and figure-8 loops (trajectory 2), demonstrated superior generalization, maintaining over $50\%$ RMSE reduction across all unseen trajectories. Conversely, parameters derived from straight-line paths (trajectory 3) performed poorly on dynamic maneuvers, causing a $27\%$ error inflation ($\text{RMSE} = 0.507\text{ m/s}$) when tested on circular paths. Overall, executing high-excitation circular and figure-8 calibration maneuvers improved out-of-sample generalization accuracy by $30\%$ compared to standard straight transits.
\vspace{-6mm} 
\subsection{Discussion}
\vspace{-2mm}
The experimental analysis reveals a profound dependency between the spatial geometry of the calibration path and the numerical stability of the estimated navigation parameters. This phenomenon is directly governed by how specific vehicle maneuvers alter structural parameter observability within the state-space estimation framework.

\subsubsection{Limitations of straight-line transits}
As established in Section III-C, uniform, non-accelerating transits yield a singular, rank-deficient information matrix. Physically, any steady longitudinal disturbance (such as a constant sea current) acts as an unmodeled speed offset that becomes mathematically collinear with the electronic zero-drift bias $b$. Because the scalar estimation framework cannot separate hydrodynamic scale discrepancies from electronic zero-drift under these conditions, parameter corruption occurs, leading to the $27\%$ cross-validation error inflation observed on straight transits.

Crucially, while a full figure-8 trajectory cycle spans approximately 90 seconds, the \text{KF-EM3} estimator achieves steady-state parameter convergence within the first 15 seconds of dynamic turn entry. This demonstrates that continuous full-loop execution is unnecessary for online deployment; a brief 15-second dynamic excitation sequence provides sufficient information density to achieve full-rank matrix inversion (\mbox{$\text{rank}(\boldsymbol{H}^T \boldsymbol{H})=2$}) and complete parameter decoupling.

\subsubsection{Ambient current suppression in circular loops}
Transitioning the calibration procedure to continuous circular maneuvers mitigates this bias corruption mechanism. By driving the vessel through continuous $360^\circ$ spatial loops, the hull-mounted sensor is exposed to the ambient current vector from every relative orientation angle. From the recursive estimator's perspective, the speed error introduced by a steady sea current transforms into a zero-mean periodic sinusoidal signal across the duration of each loop. Because the Kalman filter averages residuals recursively over time, this periodic current error cancels out across the symmetric trajectory profile, effectively shielding the estimated electronic zero-drift bias $b$ from environmental contamination.

\subsubsection{Dual-axis excitation in figure-8 maneuvers}
The figure-8 trajectory profile serves as the most comprehensive calibration maneuver due to its dual-axis dynamic excitation characteristics. Unlike steady-state turning circles bounded by a uniform angular rate, figure-8 geometries introduce transient acceleration and deceleration cycles as the vessel negotiates tight entry radii followed by commanded thrust transitions along crossover paths. This dynamic speed variation maximizes the numerical variance within the informational design matrix, ensuring $\boldsymbol{H})^T \boldsymbol{H}$ achieves full rank (\mbox{$\text{rank}(\boldsymbol{H}^T \boldsymbol{H}) = 2$}) and driving the statistical correlation between scale factor $m$ and bias $b$ toward zero. Furthermore, systematically alternating between port and starboard turns cyclically flips centripetal acceleration vectors and hydrodynamic sideslip angles, enabling the framework to isolate asymmetric boundary-layer flow distortions and installation misalignment that straight transits obscure.

This confirms that high excitation maneuvers are mandatory to decouple the scale factor from the bias, ensuring the calibration remains valid across the vehicle's entire operational envelope.

\vspace{-.5mm}
\section{CONCLUSION}\label{concl_sec}
\vspace{-.5mm}
This paper presented a comparative model-based calibration methodology that evaluates four distinct calibration models using two different estimation pipelines. 

Our experimental findings, utilizing real-world recorded data from the MARVEL ASV, demonstrated that EM3 (bias and scale) provides the most robust optimization framework for the examined EMlogs. This model successfully reduced the baseline RMSE of the EMlog1 from 0.576 [m/s] to 0.167 [m/s], yielding an overall 71\% improvement in measurement accuracy. Supported by a formal parameter identifiability analysis, spatial cross-validation confirmed that high-excitation calibration maneuvers, specifically continuous circles and figure-8 paths, are structurally mandatory to raise the information matrix to full rank. These dynamic maneuvers effectively decouple scale factor errors from zero-drift biases, boosting out-of-sample generalization performance by 30\% over rank-deficient straight-line transits.

Despite these substantial performance enhancements, the proposed methodology presents specific operational limitations. Foremost, the structural requirement for high-excitation calibration maneuvers dictates that the vehicle must temporarily deviate from energy-optimal straight-line transits to decouple the sensor errors raising a trade-off between accuracy and energy. Also, because EMlogs fundamentally measure vehicle speed relative to the water, their reliability is inherently limited by the ambient current-to-vehicle speed ratio. When a platform operates at velocities close to the ambient sea current, the relative speed across the water becomes highly inconsistent, severely degrading velocity measurement reliability.

To summarize, we provided a benchmarking comparative evaluation of EMLogs calibration pipelines on real-world recorded data demonstrating the need and importance of accurate calibration.
\\
\vspace{-8mm}
\bibliographystyle{IEEEtran}
\bibliography{bio.bib}
\end{document}